\pdfoutput=1
\documentclass[11pt]{article}

\usepackage[preprint]{acl}

\usepackage{times}
\usepackage{latexsym}
\usepackage[T1]{fontenc}
\usepackage[utf8]{inputenc}
\usepackage{microtype}
\usepackage{inconsolata}
\usepackage{graphicx}
\usepackage{booktabs}
\usepackage{amsmath}
\usepackage{xcolor}
\usepackage{xspace}
\usepackage{enumitem}
\usepackage{tikz}
\usetikzlibrary{arrows.meta, positioning, calc}
\usepackage{pifont}
\usepackage{fontawesome5}
\definecolor{cLone}{RGB}{0,114,178}
\definecolor{cLtwo}{RGB}{0,158,115}
\definecolor{cLthree}{RGB}{230,159,0}
\definecolor{cPlan}{RGB}{204,121,167}
\definecolor{cAns}{RGB}{213,94,0}
\definecolor{cInk}{RGB}{40,40,40}

\newcommand{\dataset}{CrossChapterQA\xspace}
\newcommand{\sys}{LadderRAG\xspace}

\title{Reading Between the Chapters: A Graph-Free
RAG Rebuilt With AI and Human Feedback}

\author{Dipto Sumit, Syeda Fabiayee Jahan, Farig Sadeque \\
  BRAC University, Dhaka, Bangladesh \\
  \texttt{dipto.sumit@g.bracu.ac.bd}}

\begin{document}
\maketitle

\begin{abstract}
Many questions about a book can only be answered by connecting passages chapters apart, such as a cough heard three times across a horse's life. We introduce \dataset, 1,853 such questions over 14 public-domain books. Each question lists the two to six chapters it needs (spanning half the book at the median) and the relation linking them, cites evidence from each, and has two independent reference answers graded blind. We use it to build \sys, which indexes a book as passages, summaries and book-level cards for about \$0.41 and answers a question for under a cent. Its first version failed mostly at retrieval. We redesigned it four times, steered by an LLM judge's verdicts and our analyses of failed answers: a design-level counterpart of RLAIF and RLHF in which feedback changes the architecture, not model weights. We also report the two redesigns that made answers worse. On the 191 questions its first version got wrong, the final version retrieves over twice as many required chapters on held-out books (recall 0.36 to 0.79), and ten annotators rate its answers 6.5/10 on average (764 ratings).
\end{abstract}

\section{Introduction}
\label{sec:intro}

In Anna Sewell's \textit{Black Beauty}, the narrator hears a cough three times: in a burning stable, after a hard season of hunting, and years later on a London cab stand, where it has become the sign of a ruined horse. A reader asked which cough recurs, and what it comes to mean, has to connect chapters 16, 24 and 40 and see that the three moments form a single decline. No passage answers the question on its own, and a passage that merely mentions a cough can lead an answer astray (Figure~\ref{fig:example}).

Questions like this are common whenever people read a long text closely, and they are where retrieval-augmented generation \citep{lewis2020rag} is weakest. A retriever returns the passages most similar to the question, and those tend to cluster in one place. Graph-based systems such as GraphRAG \citep{edge2024graphrag} were built for questions about a whole corpus: they extract an entity graph and summarize its communities, at the cost of many model calls per document, which later work has tried to reduce \citep{edge2024lazygraphrag, guo2024lightrag}. Trees of summaries offer another route \citep{sarthi2024raptor}. What has been missing is a way to check, question by question, whether a system found the places an answer depends on.

Long-book benchmarks already show that models struggle when evidence is spread across a novel \citep{wang2025novelqa, gupta2025novelhopqa, karpinska2024nocha}, and some attach supporting evidence to each answer \citep{wang2025novelqa}. None that we know of ensures that every question spans distant parts of the book, or records the relation that connects them, so it is hard to tell whether a wrong answer failed at retrieval or at reasoning. \dataset does both: every question lists two or more far-apart sections with evidence from each, and is typed by one of eight relations, from setup and payoff to the rationale behind the book's ordering.

We then use the benchmark to build \sys, which keeps GraphRAG's goal of answering questions about a whole book but drops the graph. \sys indexes each book once as a ladder of passages, summaries and five book-level cards in a single index. Its first version answered 1,119 sampled questions for \$0.0031 each, and an LLM judge found 83\% of the answers consistent with the references. The remaining 17\% were mostly retrieval failures. We improved the system with a loop that borrows the logic of RLAIF and RLHF \citep{bai2022constitutional, lee2024rlaif, ouyang2022instructgpt} but applies it to design rather than weights. Each version answered a development book's questions; an LLM judge flagged answers that contradicted the references; we wrote analyses of the failed answers; and the next version was designed against that feedback. No model was trained.

\begin{figure}[t]
\centering
\begin{tikzpicture}[x=0.138cm, y=1cm]
  \fill[gray!7, rounded corners=2pt] (-4.6,-1.42) rectangle (50.2,1.08);
  \draw[line width=0.8pt, gray!70] (1,0) -- (49,0);
  \foreach \c in {1,...,49} \draw[gray!55, line width=0.35pt] (\c,-0.05) -- (\c,0.05);
  \node[font=\scriptsize, gray!75!black] at (1,-0.2) {1};
  \node[font=\scriptsize, gray!75!black] at (49,-0.2) {49};
  \node[font=\scriptsize\itshape, gray!75!black, anchor=west] at (0.6,0.82) {\textit{Black Beauty}, chapters 1--49};
  \draw[cLone, line width=1.6pt] (16,0) .. controls (16,0.6) and (24,0.6) .. (24,0);
  \draw[cLone, line width=1.6pt] (24,0) .. controls (24,1.0) and (40,1.0) .. (40,0);
  \foreach \c in {16,24,40} \fill[cLone] (\c,0) circle (1.8pt);
  \node[font=\scriptsize\bfseries, text=cLone!85!black] at (16,-0.2) {16};
  \node[font=\scriptsize\bfseries, text=cLone!85!black] at (24,-0.2) {24};
  \node[font=\scriptsize\bfseries, text=cLone!85!black] at (40,-0.2) {40};
  \node[font=\tiny\itshape, text=cLone!85!black] at (16,-0.4) {fire};
  \node[font=\tiny\itshape, text=cLone!85!black] at (24,-0.4) {hunting};
  \node[font=\tiny\itshape, text=cLone!85!black] at (40,-0.4) {cab stand};
  \node[font=\scriptsize\bfseries, text=cAns!85!black, anchor=east] at (-0.4,-0.78) {v1};
  \foreach \c in {18,19,21,26,45}
    \draw[cAns!90!black, line width=0.9pt] (\c,-0.78) +(-1.7pt,-1.7pt) -- +(1.7pt,1.7pt)
                                           (\c,-0.78) +(-1.7pt,1.7pt) -- +(1.7pt,-1.7pt);
  \node[font=\scriptsize\bfseries, text=cLtwo!70!black, anchor=east] at (-0.4,-1.15) {v4b};
  \foreach \c in {18,21,43,45} \fill[cLtwo!40] (\c,-1.15) circle (1.6pt);
  \foreach \c in {16,24,40} \fill[cLtwo!75!black] (\c,-1.15) circle (2.2pt);
\end{tikzpicture}

\vspace{4pt}
\setlength{\fboxsep}{3pt}
\noindent\colorbox{cLone!9}{\parbox{\dimexpr\columnwidth-2\fboxsep\relax}{\small
\textbf{\color{cLone!80!black}Question} (\textit{recurring device}): Which cough is heard three times, and what does it come to mean?}}\par\vspace{2pt}
\noindent\colorbox{cAns!9}{\parbox{\dimexpr\columnwidth-2\fboxsep\relax}{\small
\textbf{\color{cAns!85!black}\ding{55}\,v1}, 0 of 3 chapters found: ``The cough is Jerry's, heard thrice in Chapter 45 [\ldots] a marker of declining health.'' \emph{Judged completely different.}}}\par\vspace{2pt}
\noindent\colorbox{cLtwo!11}{\parbox{\dimexpr\columnwidth-2\fboxsep\relax}{\small
\textbf{\color{cLtwo!70!black}\ding{51}\,v4b}, 3 of 3 found: ``The cough is Ginger's: the narrator hears her coughing during the stable fire, again as a `short cough' after hard hunting rides, and later [\ldots] her `frequent cough' [\ldots] in the cab-rank.'' \emph{Human scores 9.5, 9.5, 9.5, 9.25.}}}
\caption{A \dataset question about \textit{Black Beauty}. The thread joins the three chapters where Ginger's cough is heard. v1 retrieved six passages from five other chapters ({\color{cAns}$\times$}), two of them about another character's cough, and answered about the wrong character. v4b retrieved all three required chapters (dark green) among its passages.}
\label{fig:example}
\end{figure}

Our contributions are:
\begin{itemize}[leftmargin=*,itemsep=1pt,topsep=2pt]
\item \dataset: 1,853 cross-chapter questions over 14 books (986k words), labelled with their required sections, relation type and per-section evidence, each with two reference answers from different frontier models graded blind (\S\ref{sec:data}).
\item \sys, a low-cost retrieval system for whole-book questions, and a documented redesign loop driven by AI and human feedback, including the two versions that made answers worse and why (\S\ref{sec:ladder}--\ref{sec:loop}).
\item A human evaluation of the final system on the 191 questions its first version failed, with 764 ratings from ten annotators and full agreement statistics (\S\ref{sec:human}).
\end{itemize}
We will release the questions, references, judgments, the code and outputs of every version, and the human ratings.\footnote{Anonymized code and data: \url{https://anonymous.4open.science/r/Ladder_rag-7840/}}

\section{Related Work}
\label{sec:related}

\paragraph{Questions over whole books.}
NarrativeQA asks questions about books and films, written from their summaries \citep{kocisky2018narrativeqa}. NovelQA includes long-dependency questions over novels longer than 200k tokens and gives evidence for its answers \citep{wang2025novelqa}, NovelHopQA varies the number of reasoning hops over novel excerpts \citep{gupta2025novelhopqa}, NoCha pairs true and false claims about recent novels \citep{karpinska2024nocha}, and DetectiveQA and $\infty$Bench probe long-context reasoning \citep{xu2024detectiveqa, zhang2024infinitebench}. \citet{xiang2025graphragbench} compare graph and vector retrieval, partly on novels. \dataset differs in two ways: every question is built to span distant sections and lists them, and every question is typed by the relation that joins them. Retrieval can therefore be scored apart from the answer, and results can be broken down by the kind of connection a question asks for.

\paragraph{Retrieval over long documents.}
Beyond flat chunk retrieval, systems add structure. RAPTOR builds a tree of recursive summaries \citep{sarthi2024raptor}; GraphRAG, its lighter variants and HippoRAG build entity graphs \citep{edge2024graphrag, edge2024lazygraphrag, guo2024lightrag, gutierrez2024hipporag}; contextual retrieval prefixes each chunk with a sentence that places it in its document \citep{anthropic2024contextual}. \sys combines summary tiers and contextual prefixes with book-level cards and with section structure read from the text itself, and builds no graph.

\paragraph{Improving systems from feedback.}
RLHF and RLAIF train models on human or AI preferences \citep{ouyang2022instructgpt, bai2022constitutional, lee2024rlaif}, and both have been applied to parts of RAG pipelines \citep{zhang2025ragreward, nguyen2024rewardrag}. Another line improves pipelines without training: DSPy tunes prompts against a metric \citep{khattab2024dspy}, TextGrad passes textual critiques back through a pipeline \citep{yuksekgonul2024textgrad}, AutoRAG searches over RAG modules \citep{kim2024autorag}, and ADAS and AFlow search over agent designs \citep{hu2025adas, zhang2025aflow}. These methods optimize automatically against a score. Our loop changes the architecture by hand, combines an LLM judge \citep{zheng2023judge} with human error analyses, and records what each change did, including the regressions. \citet{shankar2024validators} show how people refine their criteria while checking LLM evaluators; we bring the same attention to the system under evaluation.

\section{\dataset}
\label{sec:data}

\paragraph{Books.}
We use 14 public-domain English books: \textit{Black Beauty}, \textit{The Call of the Wild}, \textit{A Christmas Carol}, \textit{The Memoirs} and \textit{The Return of Sherlock Holmes}, \textit{The Hound of the Baskervilles}, \textit{The Invisible Man}, \textit{Little Women}, \textit{Persuasion}, \textit{The Prisoner of Zenda}, \textit{The Railway Children}, \textit{The Sign of the Four}, \textit{A Study in Scarlet} and \textit{Tarzan of the Apes}. Together they hold 986k words in 290 sections: chapters, the five staves of \textit{A Christmas Carol}, and the stories of the two Holmes collections. We cleaned each text and marked every section with an explicit header, so that a label such as ``Chapter~27'' or ``Stave~III'' points to an exact span (Appendix~\ref{app:books}).

\paragraph{Questions.}
Every question can be answered only by combining two or more places in the book that are far apart, and its answer depends on how those places relate. The questions were written by an LLM, Claude Opus~5.5, working through each book chapter by chapter in an agentic session and writing questions from the text it was reading. Each question carries a list of required sections, a relation type and supporting evidence from every required section, mostly as verbatim quotations. Automatic checks confirm that every question requires at least two sections, that the first and last are at least two sections apart (one stave apart in \textit{A Christmas Carol}, whose five staves run to about 6,000 words each), that no question names a chapter, stave or story, since that would hand the retrieval target to the system, and that no question duplicates another. We also reviewed the wording of all 254 \textit{Black Beauty} questions.

\begin{figure*}[t]
\centering
\includegraphics[width=\textwidth]{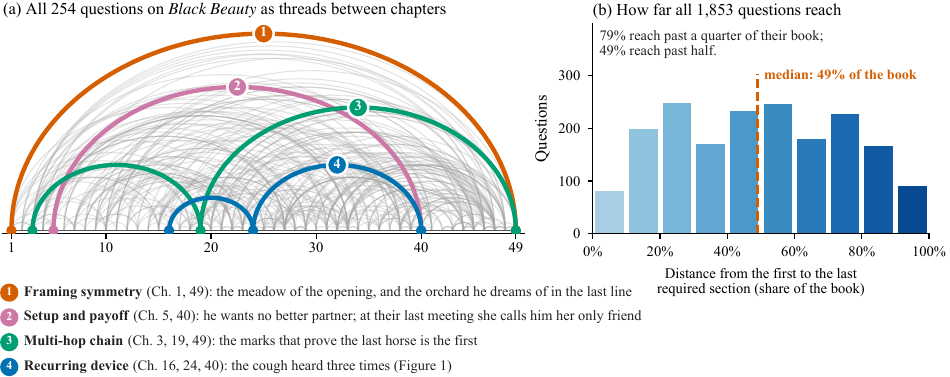}
\caption{\dataset questions as threads. (a) Each grey arc joins two chapters that one \textit{Black Beauty} question requires; four questions of different relation types are highlighted, one of them the cough of Figure~\ref{fig:example}. (b) For all 1,853 questions, the distance between the first and last required section as a share of the book: answering usually means connecting parts of a book that lie far apart.}
\label{fig:threads}
\end{figure*}

Table~\ref{tab:types} lists the eight relation types. Most questions need three sections (53\%), 36\% need two and 11\% need four to six. The first and last required sections lie a median of 9 sections apart, which at the median is 49\% of the book. Questions average 17 words. Figure~\ref{fig:threads} draws the questions as threads between the sections they join: 79\% reach past a quarter of their book and half reach past its middle, so the evidence is rarely on neighbouring pages.

\begin{table}[t]
\centering
\small
\setlength{\tabcolsep}{4pt}
\begin{tabular}{@{}p{0.70\columnwidth}r@{}}
\toprule
\textbf{Relation type}: what the answer must connect & \textbf{n} \\
\midrule
\textit{Recurring device}: a thing, phrase or event that returns, and what changes between returns & 360 \\
\textit{Setup and payoff}: an early detail and the later scene that pays it off & 317 \\
\textit{Arc}: how something develops across three or more stages & 240 \\
\textit{Local element, global function}: a small detail and its larger role elsewhere & 209 \\
\textit{Document-wide technique}: a narrative technique and its instances across the book & 201 \\
\textit{Framing symmetry}: two scenes that mirror or contrast & 187 \\
\textit{Multi-hop chain}: a chain in which each link leads to the next & 170 \\
\textit{Ordering rationale}: why the book places one part before another & 169 \\
\bottomrule
\end{tabular}
\caption{Relation types in \dataset (1,853 questions).}
\label{tab:types}
\end{table}

\paragraph{Reference answers.}
Each question has two reference answers, written independently. Claude Opus~5.5 wrote one while composing the question, with the evidence in view. GPT-5.7 Sol wrote the other from the question, its required sections and the book text, without seeing the first. The two styles differ (31 and 51 words on average), so before grading we normalized typography, removed every field that could identify the source, and shuffled each pair under neutral labels, balancing the positions within each book. A third model, OpenAI's GPT-6 Astra, then graded every answer blind, using only the text of the required sections and a 100-point rubric: accuracy (30), coverage of the required sections (20), the relation between them (20), completeness (15), specificity (10) and discipline (5), with caps for invented events and for denying the question's premise (Appendix~\ref{app:rubric100}). Neither source dominated. Opus's answer won 578 questions and GPT's 561, and 714 were ties (totals within two points); both averaged 91.4. We take the winning answer, or both answers on a tie, as the gold reference. Gold answers average 97.2, and 88\% score at least 90.

\section{\sys}
\label{sec:ladder}

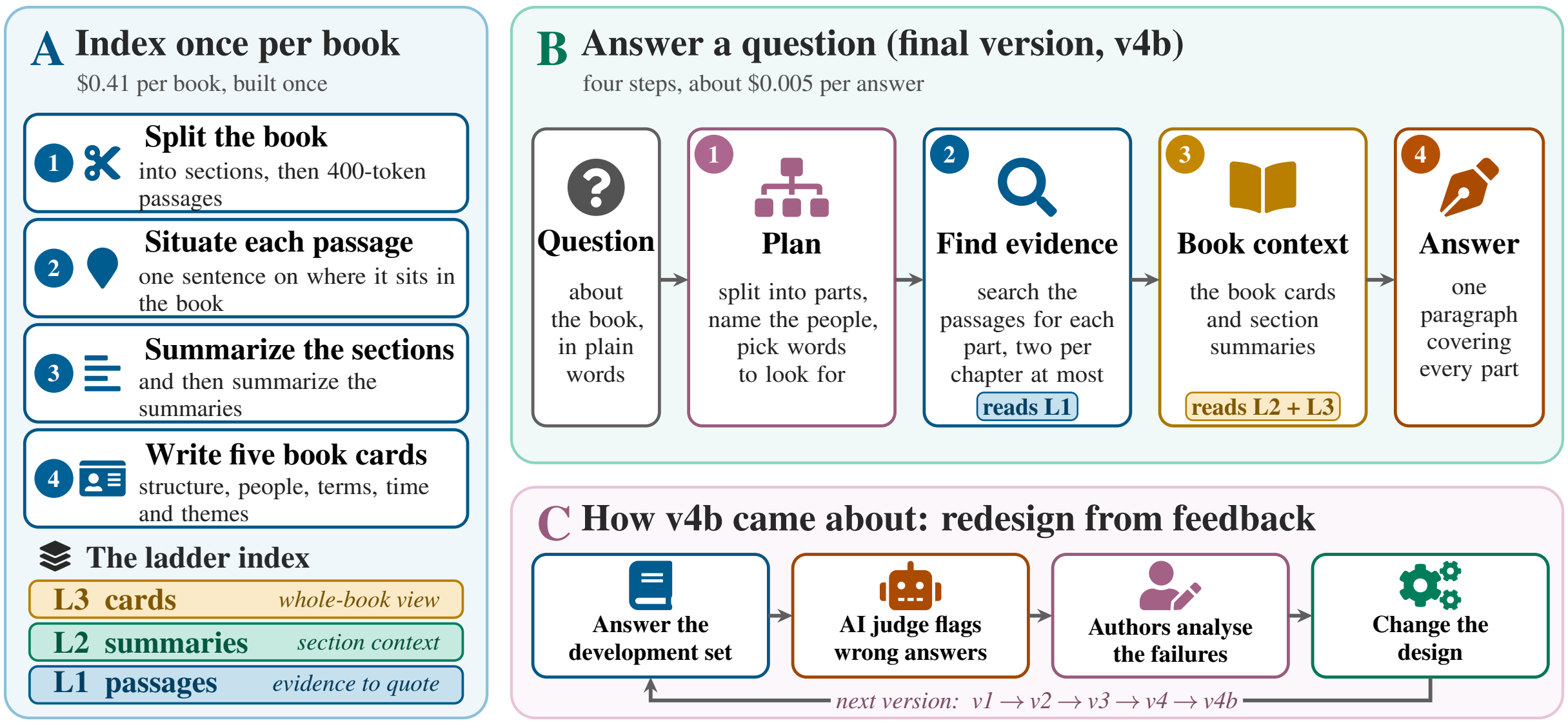
\begin{figure*}[t]
\centering
\resizebox{\textwidth}{!}{%
\begin{tikzpicture}[
  font=\footnotesize,
  panel/.style={rounded corners=7pt, fill=#1!6, draw=#1!45, line width=0.8pt},
  card/.style={rounded corners=4pt, fill=white, draw=#1!75!black, line width=0.9pt},
  num/.style={circle, fill=#1, text=white, font=\scriptsize\bfseries, inner sep=0pt, minimum size=0.4cm},
  arr/.style={-{Stealth[length=4.5pt, width=4pt]}, line width=1.1pt, cInk!75},
  chip/.style={rounded corners=2pt, fill=#1!22, draw=#1!80!black, line width=0.5pt,
               font=\scriptsize\bfseries, text=#1!55!black, inner sep=1.6pt},
  letter/.style={font=\Large\bfseries, text=#1!75!black, anchor=west},
  head/.style={font=\normalsize\bfseries, anchor=west, text=cInk},
  sub/.style={font=\scriptsize, anchor=west, text=gray!60!black},
]
\draw[panel=cLone] (0,0) rectangle (4.95,7.3);
\node[letter=cLone] at (0.12,6.88) {A};
\node[head] at (0.58,6.9) {Index once per book};
\node[sub] at (0.6,6.5) {\$0.41 per book, built once};
\foreach \yy/\nn/\ic/\ti/\de in {
    5.7/1/\faCut/{Split the book}/{into sections, then 400-token passages},
    4.62/2/\faMapMarker/{Situate each passage}/{one sentence on where it sits in the book},
    3.54/3/\faAlignLeft/{Summarize the sections}/{and then summarize the summaries},
    2.46/4/\faIdCard/{Write five book cards}/{structure, people, terms, time and themes}} {
  \draw[card=cLone] (0.2,\yy-0.5) rectangle (4.75,\yy+0.5);
  \node[num=cLone!85!black] at (0.5,\yy) {\nn};
  \node[text=cLone!80!black, font=\large] at (1.0,\yy) {\ic};
  \node[anchor=west, font=\footnotesize\bfseries] at (1.3,\yy+0.25) {\ti};
  \node[anchor=north west, font=\scriptsize, text=gray!25!black, text width=3.3cm, align=left,
        inner sep=1pt, execute at begin node={\hyphenpenalty=10000\exhyphenpenalty=10000}]
        at (1.33,\yy+0.05) {\de};
}
\node[anchor=west, font=\footnotesize\bfseries, text=cInk] at (0.22,1.66) {\faLayerGroup\ \ The ladder index};
\foreach \yy/\col/\lv/\nm/\rd in {1.22/cLthree/L3/cards/{whole-book view},
                                  0.78/cLtwo/L2/summaries/{section context},
                                  0.34/cLone/L1/passages/{evidence to quote}} {
  \draw[rounded corners=2.5pt, fill=\col!22, draw=\col!80!black, line width=0.7pt] (0.25,\yy-0.19) rectangle (4.7,\yy+0.19);
  \node[anchor=west, font=\footnotesize\bfseries, text=\col!55!black] at (0.36,\yy) {\lv\ \ \nm};
  \node[anchor=east, font=\scriptsize\itshape, text=\col!50!black] at (4.6,\yy) {\rd};
}
\draw[panel=cLtwo] (5.2,2.62) rectangle (16,7.3);
\node[letter=cLtwo] at (5.32,6.88) {B};
\node[head] at (5.78,6.9) {Answer a question (final version, v4b)};
\node[sub] at (5.8,6.5) {four steps, about \$0.005 per answer};
\draw[card=gray] (5.42,3.0) rectangle (6.72,6.05);
\node[text=gray!65!black, font=\LARGE] at (6.07,5.45) {\faQuestionCircle};
\node[font=\footnotesize\bfseries] at (6.07,4.88) {Question};
\node[font=\scriptsize, align=center, text width=1.12cm, text=gray!25!black, anchor=north] at (6.07,4.6) {about the book, in plain words};
\foreach \xx/\ww/\tw/\col/\nn/\ic/\ti/\de in {
    7.02/2.12/1.85cm/cPlan/1/\faSitemap/{Plan}/{split into parts, name the people, pick words to look for},
    9.44/2.12/1.85cm/cLone/2/\faSearch/{Find evidence}/{search the passages for each part, two per chapter at most},
    11.86/2.12/1.85cm/cLthree/3/\faBookOpen/{Book context}/{the book cards and section summaries},
    14.28/1.52/1.3cm/cAns/4/\faPenNib/{Answer}/{one paragraph covering every part}} {
  \draw[card=\col] (\xx,3.0) rectangle (\xx+\ww,6.05);
  \node[num=\col!85!black] at (\xx+0.26,5.79) {\nn};
  \node[text=\col!80!black, font=\LARGE] at (\xx+\ww/2,5.45) {\ic};
  \node[font=\footnotesize\bfseries] at (\xx+\ww/2,4.88) {\ti};
  \node[font=\scriptsize, align=center, text width=\tw, text=gray!25!black, anchor=north] at (\xx+\ww/2,4.6) {\de};
}
\node[chip=cLone] at (10.5,3.2) {reads L1};
\node[chip=cLthree] at (12.92,3.2) {reads L2 + L3};
\foreach \aa/\bb in {6.72/7.02, 9.14/9.44, 11.56/11.86, 13.98/14.28}
  \draw[arr] (\aa,4.5) -- (\bb,4.5);
\draw[panel=cPlan] (5.2,0) rectangle (16,2.37);
\node[letter=cPlan] at (5.32,2.0) {C};
\node[head] at (5.78,2.02) {How v4b came about: redesign from feedback};
\foreach \xx/\col/\ic/\ti in {
    5.42/cLone/\faBook/{Answer the\\development set},
    8.09/cAns/\faRobot/{AI judge flags\\wrong answers},
    10.76/cPlan/\faUserEdit/{Authors analyse\\the failures},
    13.43/cLtwo/\faCogs/{Change the\\design}} {
  \draw[card=\col] (\xx,0.42) rectangle (\xx+2.42,1.68);
  \node[text=\col!80!black, font=\Large] at (\xx+1.21,1.36) {\ic};
  \node[font=\scriptsize\bfseries, align=center] at (\xx+1.21,0.8) {\ti};
}
\foreach \aa/\bb in {7.84/8.09, 10.51/10.76, 13.18/13.43}
  \draw[arr] (\aa,1.05) -- (\bb,1.05);
\draw[arr] (14.64,0.42) -- (14.64,0.18) -- (6.63,0.18) -- (6.63,0.42);
\node[font=\scriptsize\itshape, fill=cPlan!6, inner sep=1.5pt, text=cPlan!60!black] at (10.6,0.18) {next version:\ \ v1\,$\rightarrow$\,v2\,$\rightarrow$\,v3\,$\rightarrow$\,v4\,$\rightarrow$\,v4b};
\end{tikzpicture}}
\caption{\sys at a glance. \textbf{(A)}~Each book is indexed once, in four steps, into a ladder with three levels: passages (L1), section summaries (L2) and five book-level cards (L3). \textbf{(B)}~The final version answers in four steps: it plans the question, finds evidence for each part among the passages, adds the book-level context, and answers in a single call (retrying with more passages if a part is missing). The coloured tags show which level of the ladder each step reads. \textbf{(C)}~How v4b came about: each version answered the development questions, an AI judge flagged wrong answers, we analysed the failures, and the design changed (\S\ref{sec:loop}).}
\label{fig:system}
\end{figure*}

\paragraph{Building the ladder.}
\sys indexes a book once. It reads the sections from their headers, with no model calls, and splits each section into chunks of about 400 tokens with 50 tokens of overlap, never crossing a section boundary. A model then writes one sentence placing each chunk in the book, in one batched call per section \citep{anthropic2024contextual}. Next it summarizes each section (splitting sections longer than 12 chunks) and adds upper tiers of summaries over groups of three, as in RAPTOR \citep{sarthi2024raptor}. From the section summaries it writes five book-level cards: a structural card that includes a one-line map of every chapter, and cards for characters, terms, chronology and recurring themes. Passages (L1), summaries (L2) and cards (L3) share one index, searched with dense vectors (bge-small; \citealp{xiao2024cpack}) and BM25 \citep{robertson2009bm25} fused by reciprocal rank \citep{cormack2009rrf}. Every indexing call uses gpt-5-mini. Building the 14 ladders took 1,196 calls and cost \$5.78 as billed, including retries and failed calls: \$0.41 per book.

\paragraph{Answering (v1).}
A router (gpt-5-nano) labels each question LOCAL, MID or GLOBAL and picks extra cards. LOCAL retrieves six passages; MID retrieves three summaries and then four passages from the sections they cover; GLOBAL retrieves two upper-tier summaries and no passages. The structural card is always in the prompt, as a cached prefix. gpt-5-mini answers in one paragraph of about 120 words or fewer and cites the sections it uses.

\paragraph{Where v1 fails.}
We sampled 80 questions per book, stratified by relation type (79 for \textit{A Christmas Carol}, which has 79), and v1 answered all 1,119 for \$0.0031 each. Word-overlap metrics say little on this task: ROUGE-L \citep{lin2004rouge} between v1's answers and the references is 0.16, but between the two references themselves, both graded correct, it is only 0.20 (Appendix~\ref{app:metrics}). We therefore asked an LLM judge, Claude Sonnet~5.5, a narrower question about each answer: is it \emph{completely different} from the reference, that is, does its main point contradict the reference or concern other events or characters? Differences in wording, missing or extra detail, partial answers and refusals do not count. The judge flagged 191 answers (17.1\%) and found the other 928 consistent with the references.

The flagged answers are mostly retrieval failures. We measure \emph{location recall}, the share of a question's required sections that appear among the retrieved passages (0 when no passages are retrieved). It averages 0.34 for the flagged answers and 0.46 for the rest (Mann--Whitney $p<10^{-4}$), and 36\% of the flagged answers retrieved none of their required sections. The router sent 55\% of all questions to LOCAL, whose six passages come from a single similarity ranking, so near-duplicate passages from one chapter crowd out the other end of the link (Figure~\ref{fig:example}). Failure rates also follow the relation types. They are highest for local elements with a global function (24\%) and for setup and payoff (23\%), whose answers hinge on two specific scenes, and lowest for document-wide techniques, arcs and ordering (10--11\%), which the book-level cards are built to cover. In 127 answers (11\%) v1 said that the retrieved passages did not contain the information. Appendix~\ref{app:v1} breaks these failures down further.

\section{Redesign from AI and Human Feedback}
\label{sec:loop}

\begin{figure}[t]
\centering
\includegraphics[width=\columnwidth]{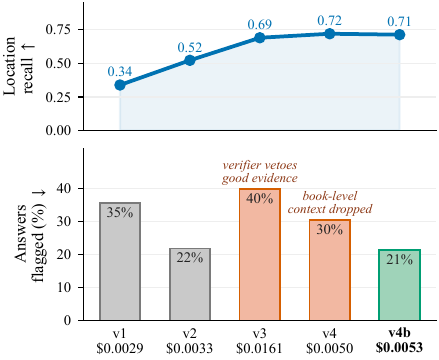}
\caption{The redesign trajectory on the development book. Retrieval (top) improved almost every round, but answers (bottom, share flagged by the AI judge) did not follow: v3 and v4 retrieved more and answered worse. Below each version: cost per answer.}
\label{fig:versions}
\end{figure}

We improved \sys on a single development book, \textit{Black Beauty}, which both authors know well enough to judge an answer without rereading. Every version answered the same 80 sampled questions from the question text alone. Three signals shaped the next design:
\begin{itemize}[leftmargin=*,itemsep=1pt,topsep=2pt]
\item \textbf{AI feedback.} gpt-oss-120b \citep{openai2025gptoss}, at high reasoning effort, applied the completely-different criterion to every answer. Re-judging identical answers moved the count by two, so differences of about two questions in 80 are noise.
\item \textbf{Human feedback.} We wrote analyses of the failed answers to 13 annotated questions, classifying each error and stating what the system should have done.
\item \textbf{Diagnostics.} Location recall, questions with no required section retrieved, refusals, and cost.
\end{itemize}
We proposed each change from these signals, implemented it with the help of an AI coding assistant, and kept it only after reviewing its results. No weights changed, and no answer was chosen by its judge score: each version answered once and was judged afterwards. In RLAIF, an AI judge's preferences update a model's weights; here the judge's verdicts, together with our written analyses, update the system's design, and each update is a code change that a person reviews. Table~\ref{tab:versions} follows the five versions, and Figure~\ref{fig:versions} plots their trajectory.

\begin{table*}[t]
\centering
\footnotesize
\setlength{\tabcolsep}{3.2pt}
\renewcommand{\arraystretch}{1.12}
\begin{tabular}{@{}l p{4.95cm} p{4.0cm} ccccc@{}}
\toprule
& \textbf{Feedback that led to the version} & \textbf{Design change} & \textbf{Flagged}\,$\downarrow$ & \textbf{Rev.}\,$\uparrow$ & \textbf{Recall}\,$\uparrow$ & \textbf{None}\,$\downarrow$ & \textbf{\$/ans.} \\
\midrule
v1 & (first version) & router picks one route; one similarity ranking per route & 28/79 & 1/13 & 0.34 & 37 & 0.0029 \\
v2 & \textit{AI judge and diagnostics:} questions linking distant chapters routed as local; passages crowded into one chapter; two-part questions searched as one string & planner routes by the scope an answer needs; up to 3 sub-queries fused by RRF; at most 2 passages per chapter & 17/78 & 5--7/13 & 0.52 & 16 & 0.0033 \\
v3 & \textit{Our review of 12 answers:} a scene with the right words but the wrong event; first and last swapped; another character's story credited to the narrator; categories merged; ``three times'' answered from one chapter & constraint planner; stemmed BM25, rare-word lookup, position masks; a verifier that can reject passages; ``not found'' allowed & 31/78 & 6/13 & 0.69 & 4 & 0.0161 \\
v4 & \textit{AI judge:} the verifier vetoed correct evidence. \textit{Our review of 13 answers:} wrong anchor passage (8 of 13); objects read as symbols; one hop where two were asked; characters lost across renames & literal anchors from the question's own words; names via the characters card; reserved slots per part; a checklist in one answer call; one retry with 14 passages & 24/79 & 8/13 & 0.72 & 4 & 0.0050 \\
\textbf{v4b} & \textit{AI judge:} 13 answers that v2 had right went wrong: interpretive questions without book-level context; back-references and speakers & restore v2's cards and summaries; rules for back-references and speakers & \textbf{17/80} & \textbf{8/13} & 0.71 & 7 & 0.0053 \\
\bottomrule
\end{tabular}
\caption{The redesign loop on the development book (80 \textit{Black Beauty} questions). Each row gives the feedback that prompted a version, the change made, and the result. \emph{Flagged}: answers the AI judge found completely different from the reference, over the questions it gave a verdict on (35\%, 22\%, 40\%, 30\%, 21\%). \emph{Rev.}: of the 13 questions we analysed by hand, those not flagged (v2 was judged twice). \emph{Recall}: mean location recall. \emph{None}: questions with no required section retrieved.}
\label{tab:versions}
\end{table*}

\paragraph{v2: plan before retrieving.}
The first round of feedback named three problems: questions linking distant chapters were routed by their wording, the passage slots went to one chapter, and a two-part question was searched as one string. In v2 a planner (gpt-5-nano) routes by how much of the book an answer needs, splits the question into up to three sub-queries that are searched separately and fused, and admits at most two passages per chapter. Recall rose from 0.34 to 0.52, and flagged answers fell from 35\% to 22\%. Letting the planner name chapters from the chapter map lowered recall to 0.49, so we switched that part off. Across all books, v2 also answered 114 of the 127 questions v1 had refused (22 of them only with twelve passages instead of six); the 13 that remained were exact-word misses, such as a ``blistered heel'' that the text describes as a blister on the heels.

\paragraph{What we found by hand.}
Our review of 12 failed answers found that eight were retrieval failures of specific kinds: a scene with the same vocabulary but the wrong event, a ``last meeting'' answered with the first, a story another character tells credited to the narrator, two of the mother's ``sorts of men'' merged into one, and something ``heard three times'' answered from a single chapter. The common cause was that nothing checked the evidence against the question's own constraints.

\paragraph{v3: a verifier, and a regression.}
v3 added that check as a separate step. A planner extracts each part's constraints (early or late in the book, a count, the speaker), retrieval adds stemmed BM25, rare-word lookup and position masks, a verifier tests every passage against every constraint, and the answer may say ``not found'' instead of substituting another scene. Recall rose to 0.69, but flagged answers rose to 40\% and the cost per answer to five times v2's. The verifier often rejected correct evidence because the book did not use the question's own words (it looked for a ``third sort'' of man, for example), and 16 of the 31 flagged answers contained ``not found''.

\paragraph{v4: anchors first.}
A second review, of 13 answers, sharpened the goal: correct and relevant answers at low cost, not stylistic parity with frontier models. Wrong anchor passages explained 8 of the 13 failures; other errors included reading literal objects as symbols, answering two-hop questions with one hop, confusing first and last, crediting the wrong speaker, and losing a character across renames (Darkie, Black Beauty, Black Auster, Jack, Blackie, Old Crony). v4 moves the fixes into retrieval and a soft checklist. A planner (gpt-5-mini) splits the question into parts, turns descriptions into character names with the help of the characters card, and proposes \emph{anchors}, literal words from the question such as ``little brown book''. Code discards any anchor not made of the question's own words, after we saw the planner steer search toward scenes it had guessed. Each part gets reserved passage slots, filled from exact anchor matches, rare terms, and dense plus stemmed-BM25 search, with one passage per chapter when a part asks for several instances. A single answer call receives the passages, a checklist of parts and a few rules (objects are literal; answer every part; say ``not found'' only for a part with no plausible passage), and the system retries once with 14 passages if a part is missing.\footnote{Two planned mechanisms of v4 and v4b, position masks for ``first'' and ``last'' and a first-person search for the narrator's own experience, never fired because of a pattern bug. All results reflect the code as it ran.} Recall reached 0.72 and v4 answered 8 of the 13 reviewed questions acceptably, but flagged answers rose back to 30\%. Without the cards and summaries, v4 lost the interpretive questions about why the book is arranged as it is, and it mixed up speakers and back-references.

\paragraph{v4b: anchors and the ladder.}
v4b restores the cards and summaries selected by v2's plan, and adds two rules: a later part that refers back (``that horse'') keeps the same character, and a line belongs only to the speaker the passage names. It ties v2 for the fewest flagged answers (21\%), keeps v4's recall and its 8 of 13 review questions, and costs \$0.0053 per answer. A further version, v5, added three more fixes and was flagged on 16 and 18 of 80 questions in two judgings, within noise of v4b, so we kept v4b.

\paragraph{What the loop taught us.}
Better retrieval did not guarantee better answers: v3 and v4 retrieved more of the required sections than v2 but answered worse, once because a separate step could veto good evidence and once because the book-level context was gone. The two kinds of feedback also did different jobs. The judge told us how many answers were wrong, and the written reviews told us why.

\section{Human Evaluation}
\label{sec:human}

\paragraph{Setup.}
For the final evaluation we ran v4b on the 191 questions whose v1 answers had been flagged: 178 from the 13 held-out books and the 13 \textit{Black Beauty} questions we had analysed. We re-answered only where v1 had gone wrong; the other 928 questions keep their v1 answers, which the judge had found consistent with the references. On the held-out questions, v4b's location recall is 0.79 against v1's 0.36, and only 6 of the 178 retrieved no required section, against 58 for v1. Each v4b answer here cost \$0.0066.

Ten annotators, paid for their time, scored v4b's answers. They worked in three pools. Pools A and B each had three members who know the stories well; pool A covered six books and pool B the other eight, and each member scored every answer in the pool's books independently. Pool C was a panel of four annotators from a university English department, who know all 14 books and scored every answer together, producing one score per answer. Every answer therefore received four scores, 764 in all. Annotators saw the question, the gold reference and v4b's answer, and scored v4b's answer against the book text on a scale from 1 to 10. The rubric (Appendix~\ref{app:human-rubric}) has them split the question into parts and mark each part correct, partly correct, missing or wrong. The answered parts set a ceiling of $2 + 8 \times$ (parts answered / parts asked), from which they deduct for wrong parts, minor errors, invented details, a missing or thinly stated link, vagueness and padding. A refusal counts as a missing part, and length earns nothing.

\begin{figure}[t]
\centering
\includegraphics[width=\columnwidth]{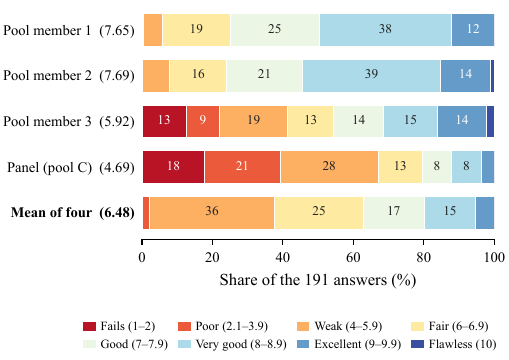}
\caption{Human scores for v4b's answers to the 191 questions v1 answered wrongly, by the rubric's score bands (mean in brackets). ``Pool member'' scores come from pool A or B, depending on the book; the panel is pool C. Numbers inside the bars are percentages.}
\label{fig:human}
\end{figure}

\paragraph{Results.}
Figure~\ref{fig:human} shows how the scores spread over the rubric's bands. Averaged over the four scores, v4b's answers to the questions v1 got wrong score 6.48, in the rubric's ``fair'' band, where every part is touched but with weaknesses. Of the 191 answers, 116 (61\%) average 6.0 or more, 71 (37\%) average 7.0 or more, the ``good'' band in which every part is answered, and only 4 fall below 4.0. The held-out books average 6.44 and the development book 7.11. Scores follow retrieval: answers whose passages covered every required section average 6.88, against 5.99 for the rest (Spearman $\rho=0.31$ between location recall and score), while answer length is unrelated to score ($\rho=0.10$; Appendix Figure~\ref{fig:recall}). Among the relation types, arcs (6.86) and setup and payoff (6.74) score highest, and framing symmetry (6.18) and ordering rationale (6.19) lowest.

\paragraph{Where v4b still falls short.}
The three lowest-scored answers show the failures that remain, and none of them is a simple retrieval miss. Asked in what order four remembered episodes of \textit{The Prisoner of Zenda} occur, v4b retrieved three of the four required chapters and still replied that the passages did not contain the narrator's list. Asked how \textit{The Sign of the Four} quotes the same author twice, it found a different book that Holmes quotes twice, a gazetteer rather than Winwood Reade. And asked why \textit{Persuasion} dismisses Dick Musgrove so harshly before Wentworth arrives, it retrieved both required chapters but explained a different dismissal, after Louisa's fall. A refusal despite the evidence, a similar scene chosen over the right one, and a ``why'' question answered with plot: the last two also remained open on the development book, and all three mark where the next round of feedback should go.

\paragraph{Agreement.}
The four scores order the answers similarly but sit at different levels: the two most lenient pool members average 7.7 and the panel 4.7. That offset pulls absolute agreement down. Krippendorff's $\alpha$ (interval; \citealp{hayes2007alpha}) over all four scores is 0.22 (95\% CI 0.16--0.28), and the single-score ICC(2,1) is 0.29 \citep{shrout1979icc}. Consistency, which ignores the offsets, is higher at ICC(3,1) = 0.46, and the mean of the four scores, the quantity we report, is reliable: ICC(2,4) = 0.62 and ICC(3,4) = 0.77. Agreement is strongest between the third pool member and the panel ($\alpha=0.67$, 95\% CI 0.58--0.74; quadratic-weighted $\kappa=0.64$, \citealp{cohen1968weighted}) and weakest between the two lenient members ($\alpha=0.28$). Fleiss' $\kappa$ \citep{fleiss1971kappa} over the rubric's eight score bands is 0.07, because band agreement counts every offset as a disagreement; Appendix~\ref{app:agreement} gives all pairwise statistics. The offset is largest on the clearest failures: on the \textit{Persuasion} answer above, the two lenient members gave 6.25 and 7.5, and the third member and the panel gave 1.0. The annotators thus rank the answers alike and differ mainly in how much they deduct for incomplete or wrong parts. Under the strictest reading, the panel's, 20\% of these answers are good or better and 39\% are poor or failing.

\section{Conclusion}
\label{sec:conclusion}

\dataset asks questions whose answers live in more than one chapter, and it records where. That record was the most useful tool we had while building a system for such questions. It showed that our first system failed at retrieval, let us measure each redesign's effect on retrieval and on answers separately, and warned us when the two came apart. The final version of \sys answers questions about a whole book without a graph for under a cent each, and on the questions its first version got wrong, human annotators now rate its answers as fair on average and good in more than a third of cases. We hope the benchmark, the version history and the ratings help others build and test systems that can read between the chapters.

\section*{Limitations}

\paragraph{Data.} The questions and the reference answers were written by LLMs. Automatic checks guard the structure of every question, and we reviewed the wording of the \textit{Black Beauty} questions, but some questions may admit valid answers that neither reference gives, and the gold answer is the better of two model answers rather than a human one. The books are English classics from the nineteenth and early twentieth centuries and are almost certainly in the pretraining data of the models we use, so some answers may draw on memory rather than on retrieved text.

\paragraph{Development and evaluation.} We developed every version on \textit{Black Beauty} and report it separately; the held-out evidence comes from the other 13 books. The screen of v1's answers rests on one LLM judge, and both judges give a coarse yes-or-no verdict; re-judging the development book moved the count by about two questions in 80. We applied v4b to the 191 questions where v1 disagreed with the references and kept v1's answers where it already agreed. Running v4b on the full sample is a natural next step to confirm that it keeps those answers; on the development book, where every version answered all 80 questions, v4b disagreed with the references less often than v1 (21\% against 35\%).

\paragraph{Human study.} Annotators knew which answer came from the system and could see the reference while scoring. Absolute agreement is low because the annotators differ in leniency, so we report the mean of four scores alongside every agreement statistic. The panel's scores are joint decisions rather than independent ones. We did not collect human scores for v1 on the same questions.

\paragraph{Comparisons.} We do not compare \sys head to head with GraphRAG, other structured retrieval systems or long-context models; our cost figures are measured on our own runs and are not matched comparisons.

\section*{Ethics Statement}

\paragraph{Data and licences.} The 14 books were published between 1817 and 1912 and are in the public domain. We took the texts from Project Gutenberg and other public-domain editions, and Appendix~\ref{app:books} credits their authors. We will release \dataset, with its questions, evidence, reference answers, grades and human ratings, under CC BY 4.0, and our code under the MIT licence, for research on question answering and retrieval over long documents. We used the proprietary OpenAI and Anthropic models through their providers' services and within their terms of use, and the open models under their licences: bge-small-en-v1.5 (MIT), gpt-oss-120b (Apache 2.0), roberta-large (MIT) and all-mpnet-base-v2 (Apache 2.0). The questions and reference answers were generated with LLMs and are labelled as such in the release.

\paragraph{Personal information and offensive content.} The data concerns the events of the books and contains no personal information, and the human ratings are released without the annotators' names. The books reflect the attitudes of their time and include dated and offensive portrayals, for example of Native Americans and Mormons in \textit{A Study in Scarlet}, of Africans in \textit{Tarzan of the Apes} and of an Andaman Islander in \textit{The Sign of the Four}. Questions, evidence and references about these passages sometimes repeat the books' terms. We keep the texts unaltered as historical documents.

\paragraph{Annotators.} We recruited the ten annotators through our academic and personal networks: four members of a university English department (pool C) and six readers who know the books well (pools A and B). All ten live in Bangladesh, and we collected no other demographic information. They were compensated fairly for their time, were told how their scores would be used, and agreed to take part. Appendix~\ref{app:human-rubric} gives their instructions. The task, scoring system answers about public-domain fiction, collected no personal data and carried no foreseeable risk, so it was not submitted to an ethics review board.

\paragraph{Risks.} The gold references are written by models, so a system tuned to them can inherit their errors; the blind grading, the second independent reference and the human evaluation reduce this risk but do not remove it. \sys can give fluent but wrong answers about a book, so it should support reading rather than replace it; its answers cite the sections they draw on, so readers can check them. Once released, the questions may leak into training data and inflate later results on the benchmark. The experiments used little compute (Appendix~\ref{app:models}).

\paragraph{AI assistance.} Besides the models studied in this paper, we used AI assistants to write code and to help draft and edit the paper and its figures. The authors checked every claim and number and take full responsibility for the content.

\bibliography{references}

\appendix

\section{Books and Sections}
\label{app:books}

\begin{table*}[t]
\centering
\small
\begin{tabular}{@{}lrrrrrc@{}}
\toprule
\textbf{Book} & \textbf{Words} & \textbf{Sections} & \textbf{Questions} & \textbf{Sampled} & \textbf{v1 flagged} & \textbf{Human mean} \\
\midrule
Black Beauty (development) & 59,886 & 49 & 254 & 80 & 13 & 7.11 \\
The Call of the Wild & 31,833 & 7 & 88 & 80 & 9 & 6.33 \\
A Christmas Carol & 29,975 & 5 & 79 & 79 & 8 & 7.77 \\
The Memoirs of Sherlock Holmes & 87,063 & 11 & 120 & 80 & 25 & 6.54 \\
The Return of Sherlock Holmes & 112,336 & 13 & 122 & 80 & 13 & 6.49 \\
The Hound of the Baskervilles & 59,193 & 15 & 130 & 80 & 6 & 6.99 \\
The Invisible Man & 48,469 & 29 & 134 & 80 & 5 & 6.34 \\
Little Women & 186,127 & 47 & 143 & 80 & 21 & 6.10 \\
Persuasion & 86,419 & 24 & 131 & 80 & 16 & 6.49 \\
The Prisoner of Zenda & 53,666 & 22 & 131 & 80 & 18 & 6.08 \\
The Railway Children & 59,377 & 14 & 138 & 80 & 10 & 5.64 \\
The Sign of the Four & 43,077 & 12 & 125 & 80 & 12 & 6.32 \\
A Study in Scarlet & 43,416 & 14 & 126 & 80 & 14 & 6.63 \\
Tarzan of the Apes & 85,523 & 28 & 132 & 80 & 21 & 6.62 \\
\midrule
Total & 986,360 & 290 & 1,853 & 1,119 & 191 & 6.48 \\
\bottomrule
\end{tabular}
\caption{The 14 books. \emph{Sampled}: questions answered by v1. \emph{v1 flagged}: v1 answers judged completely different, which v4b then answered for human scoring. \emph{Human mean}: mean of the four scores over those answers.}
\label{tab:books}
\end{table*}

\begin{figure*}[t]
\centering
\includegraphics[width=\textwidth]{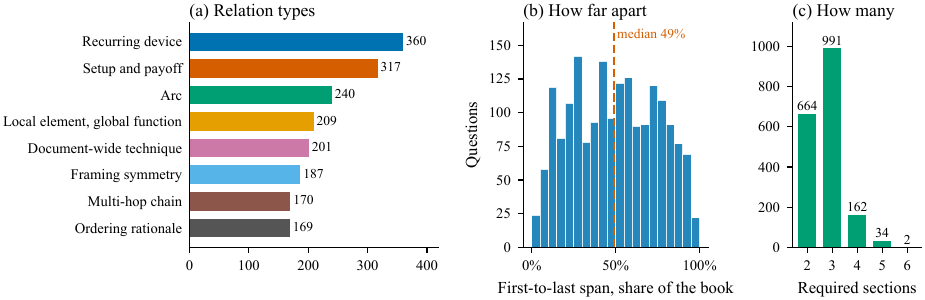}
\caption{What \dataset asks. (a) Questions per relation type. (b) Distance between the first and last required section, as a share of the book. (c) Number of required sections per question.}
\label{fig:dataset}
\end{figure*}

Table~\ref{tab:books} gives per-book statistics and Figure~\ref{fig:dataset} the make-up of the questions.

 Every cleaned text marks each section with an explicit header. For indexing we also removed the table of contents and illustration captions from \textit{A Christmas Carol} and the trailing Project Gutenberg licence from \textit{Persuasion}. The two Holmes collections are divided into stories rather than chapters. Question labels use the book's own numbering style (Arabic or Roman), and every label in the dataset resolves to exactly one marked section.

\paragraph{Sources and coverage.} We took the texts from Project Gutenberg and other public-domain editions. All 14 books are English fiction by seven British and three American authors, published between 1817 and 1912: Jane Austen (\textit{Persuasion}, 1817), Charles Dickens (\textit{A Christmas Carol}, 1843), Louisa May Alcott (\textit{Little Women}, 1868--69), Anna Sewell (\textit{Black Beauty}, 1877), Arthur Conan Doyle (\textit{A Study in Scarlet}, 1887; \textit{The Sign of the Four}, 1890; \textit{The Memoirs of Sherlock Holmes}, 1894; \textit{The Hound of the Baskervilles}, 1902; \textit{The Return of Sherlock Holmes}, 1905), Anthony Hope (\textit{The Prisoner of Zenda}, 1894), H.~G.~Wells (\textit{The Invisible Man}, 1897), Jack London (\textit{The Call of the Wild}, 1903), E.~Nesbit (\textit{The Railway Children}, 1906) and Edgar Rice Burroughs (\textit{Tarzan of the Apes}, 1912). They span detective fiction, adventure, the novel of manners, children's fiction, animal stories, a ghost story and scientific romance. The questions and answers are also in English, and the release documents every field of the data.

\section{Where v1 Fails, in Detail}
\label{app:v1}

\begin{figure*}[t]
\centering
\includegraphics[width=\textwidth]{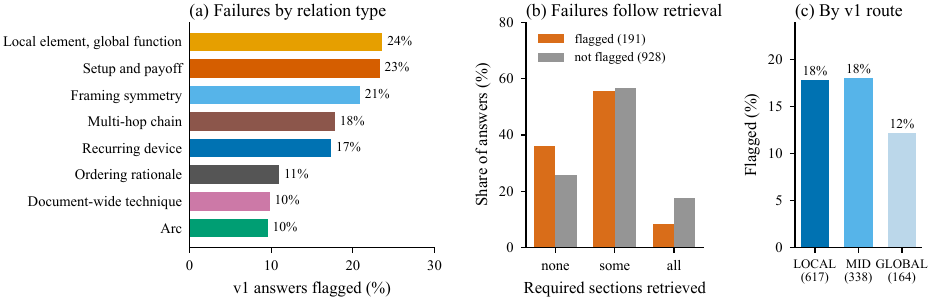}
\caption{v1 on the 1,119 sampled questions. (a) Share of answers the screening judge flagged, by relation type. (b) How many of a question's required sections v1 retrieved, for flagged and unflagged answers. (c) Flag rate by the route v1's router chose (number of questions in brackets).}
\label{fig:v1}
\end{figure*}

Figure~\ref{fig:v1} breaks down v1's failures. Questions that hinge on two specific scenes fail most often (a). Flagged answers are more likely to have retrieved none of their required sections and less likely to have retrieved all of them (b). The GLOBAL route, which answers from the book-level cards and summaries alone, fails least (c), even though it retrieves no passages.

\section{Example Questions}
\label{app:examples}

\begin{table*}[t]
\centering
\small
\setlength{\tabcolsep}{4pt}
\begin{tabular}{@{}p{3.1cm}p{2.9cm}p{6.6cm}p{2.3cm}@{}}
\toprule
\textbf{Relation type} & \textbf{Book} & \textbf{Question} & \textbf{Requires} \\
\midrule
Setup and payoff & Black Beauty & Which wish of the worn-out mare does the narrator afterwards repeat as his own? & Ch.\ 40, 47 \\
Recurring device & The Hound of the Baskervilles & How does a black beard recur as a false clue? & Ch.\ IV, V, VII \\
Arc & The Call of the Wild & How do the dog's masters range from fair to fatal to ideal? & Ch.\ 1, 5, 6 \\
Local element, global function & The Sign of the Four & Why does the detective bother to have a green lamp removed from a boat? & Ch.\ VIII, IX, X \\
Document-wide technique & The Railway Children & How does the keeping and breaking of secrets organise the book? & Ch.\ I, VI, XI, XIII, XIV \\
Framing symmetry & The Invisible Man & How is the hidden face of the opening matched at the close? & Ch.\ I, XXVIII \\
Multi-hop chain & Tarzan of the Apes & How does Kala's death lead, step by step, to Tarzan becoming king? & Ch.\ IX, X, XI \\
Ordering rationale & Persuasion & Why is Mrs Smith's revelation held back until after the concert? & Ch.\ XVII, XX, XXI \\
\bottomrule
\end{tabular}
\caption{One \dataset question of each relation type. Questions describe scenes by their content and never name the sections they need.}
\label{tab:examples}
\end{table*}

Table~\ref{tab:examples} shows one question of each relation type. Questions describe characters and scenes by content (``the worn-out mare'', ``a black beard''), so finding the right sections is part of the task.

\section{Rubric for Grading the References}
\label{app:rubric100}

The blind judge scored every reference answer out of 100, using only the text of the question's required sections. \emph{Accuracy} (30) starts at 30 and loses 10 for each major error on something the question asks about, 4 for each minor error and 2 for each plausible detail the text does not contain. \emph{Coverage} (20) is proportional to the required sections the answer actually draws on. \emph{Relationship} (20) asks whether the answer states the connection the question's type calls for (for example, how a payoff answers its setup, or how each link of a chain leads to the next). \emph{Completeness} (15) is divided among the parts of the question. \emph{Specificity} (10) rewards concrete, identifiable detail over generic statements, and \emph{discipline} (5) deducts for outside knowledge, speculation stated as fact, and material that does not address the question. Caps override the total: 20 for answering a different question, 30 for rejecting the question's premise, and 40 for inventing an event, character or quotation. Length, naming style and tone are explicitly excluded, and a worked example on a book outside the set calibrates the scale.

\section{Human Annotation Rubric}
\label{app:human-rubric}

Annotators scored each answer from 1.0 to 10.0 with one decimal, in four steps. (1) Split the question into parts: each \textit{what}, \textit{which} or \textit{who} it asks, each item in a list, and each named thing in a comparison. (2) Mark each part Correct, Partly, Missing (including ``the book does not say'') or Wrong, and set the ceiling to $2 + 8 \times$ (parts answered / parts asked): 10.0 when all parts are answered, 6.0 for half, 2.0 for none. (3) Deduct 1.0 for each wrong part, 0.5 to 2.0 for each partly answered part, 0.5 for each minor error, 0.3 for each invented detail, 1.0 for a missing link between the places (0.5 if thin), 0.3 to 1.0 for vagueness where the book is concrete, and at most 0.5 for padding; never below 1.0. (4) Check the result against the score bands (10.0 flawless, 9.0--9.9 excellent, 8.0--8.9 very good, 7.0--7.9 good, 6.0--6.9 fair, 4.0--5.9 weak, 2.1--3.9 poor, 1.0--2.0 fails), with an adjustment of at most 0.5 that must be explained in a note. Annotators worked alone, kept the book text at hand, used no AI tools, and were told that the book is the only evidence and that length and style earn nothing.

\section{Screening Criterion and Models}
\label{app:models}

Both AI judges (Claude Sonnet~5.5 for the screen of all 1,119 v1 answers, gpt-oss-120b for every version on the development book) saw the question, the gold reference answer or answers, and the system's answer, and returned \emph{completely different} or not. An answer is completely different only if its main point contradicts the reference or is about different facts, events or characters. Differences in wording or detail, partial answers, and answers saying that the passages do not contain the information are \emph{not} completely different. When two references tied, an answer is flagged only if it differs from both. For gpt-oss-120b we capped reasoning at about 4k tokens and re-ran calls that returned no verdict.

All LLM calls made by \sys went through the OpenRouter API. The parameter counts of the proprietary models are not public; gpt-oss-120b has 117B parameters (5.1B active per token), and bge-small-en-v1.5, which runs locally, has 33M. Token counts, including the 400-token chunk size, use tiktoken's \texttt{cl100k\_base} encoding. Indexing and every answering run of \sys, including the versions we did not adopt, cost about \$14 in API fees.

\begin{table}[h]
\centering
\small
\setlength{\tabcolsep}{4pt}
\begin{tabular}{@{}p{3.5cm}p{3.3cm}@{}}
\toprule
\textbf{Stage} & \textbf{Model (reasoning effort)} \\
\midrule
Situating sentences & gpt-5-mini (minimal) \\
Summaries, all tiers & gpt-5-mini (medium) \\
Book-level cards & gpt-5-mini (high) \\
Router (v1), planner (v2) & gpt-5-nano (low) \\
Planners (v3, v4, v4b) & gpt-5-mini (low) \\
Verifier (v3), answers & gpt-5-mini (medium) \\
Embeddings & bge-small-en-v1.5 \\
Screening judge & Claude Sonnet 5.5 \\
Development judge & gpt-oss-120b (high) \\
References & Claude Opus 5.5, GPT-5.7 Sol \\
Reference grading & GPT-6 Astra \\
\bottomrule
\end{tabular}
\caption{Models used at each stage. Situating sentences for two books were made at gpt-5-mini's default effort before we set it to minimal.}
\label{tab:models}
\end{table}

\section{Agreement Details}
\label{app:agreement}

\begin{table}[h]
\centering
\small
\setlength{\tabcolsep}{4pt}
\begin{tabular}{@{}lccc@{}}
\toprule
\textbf{Pair} & \textbf{Pearson} $r$ & \textbf{Spearman} $\rho$ & \textbf{QW} $\kappa$ \\
\midrule
Member 1 -- Member 2 & 0.28 & 0.31 & 0.28 \\
Member 1 -- Member 3 & 0.35 & 0.43 & 0.26 \\
Member 1 -- Panel & 0.57 & 0.60 & 0.24 \\
Member 2 -- Member 3 & 0.36 & 0.42 & 0.24 \\
Member 2 -- Panel & 0.47 & 0.51 & 0.18 \\
Member 3 -- Panel & 0.78 & 0.79 & 0.64 \\
\bottomrule
\end{tabular}
\caption{Pairwise agreement over the 191 answers. QW $\kappa$: quadratic-weighted kappa over the rubric's eight bands.}
\label{tab:pairwise}
\end{table}

Table~\ref{tab:pairwise} gives pairwise agreement. Rank correlations exceed the weighted kappas for every pair involving a lenient pool member, which is the signature of a level offset rather than a disagreement about order. Over all four scores, Krippendorff's $\alpha$ is 0.22 on the interval scale; ICC(2,1) = 0.29, ICC(2,4) = 0.62, ICC(3,1) = 0.46 and ICC(3,4) = 0.77. Confidence intervals come from 2,000 bootstrap resamples of the 191 answers.

\begin{figure*}[t]
\centering
\includegraphics[width=\textwidth]{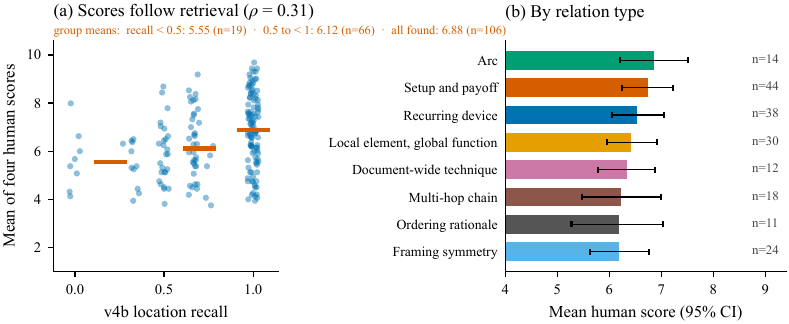}
\caption{Human scores for v4b's 191 answers. (a) Mean of the four scores against v4b's location recall (jittered); bars mark the mean for answers that found fewer than half, at least half, or all of their required sections. (b) Mean score by relation type, with bootstrap 95\% confidence intervals.}
\label{fig:recall}
\end{figure*}

Figure~\ref{fig:recall} shows the link between retrieval and human scores (a) and the spread across relation types (b). The intervals in (b) overlap widely: with 11 to 44 answers per type, the differences between types are suggestive rather than settled.

\section{Word-Overlap Metrics}
\label{app:metrics}

\begin{table}[h]
\centering
\small
\setlength{\tabcolsep}{5pt}
\begin{tabular}{@{}lcc@{}}
\toprule
\textbf{Metric} & \textbf{v1 vs.\ refs.} & \textbf{Ref.\ vs.\ ref.} \\
\midrule
ROUGE-1 & 0.26 & 0.28 \\
ROUGE-2 & 0.05 & 0.06 \\
ROUGE-L & 0.16 & 0.20 \\
BERTScore F1 & 0.08 & 0.15 \\
Embedding cosine & 0.62 & 0.54 \\
\bottomrule
\end{tabular}
\caption{Overlap between v1's 1,119 answers and the two references (mean over the references), and between the two references. ROUGE uses F-measures with stemming, BERTScore uses roberta-large with baseline rescaling \citep{zhang2020bertscore}, and the embedding is all-mpnet-base-v2.}
\label{tab:metrics}
\end{table}

Two correct answers to the same question share few words (Table~\ref{tab:metrics}), so overlap metrics cannot separate right answers from wrong ones on this task. This is why we rely on an LLM judge for screening and on human annotators for the final evaluation.

\section{v4b Answer Instructions}
\label{app:rules}

After the cards, summaries, passages and question (the same layout as v1), v4b's answer prompt ends with the following text, in which \texttt{\{checklist\}} lists the planner's parts and their anchors:

\begin{quote}
\small
CHECKLIST - the question's parts and conditions: \texttt{\{checklist\}}

Before writing, pick for each part the passage whose scene matches the question's own description (the exact object, person, place, first/last, who is speaking, which category). If two scenes are similar, use the one that matches the description, not the more general one. Objects in the question are literal objects in the story, not symbols. When a later part refers back (``that horse'', ``the same way'', ``that description''), it is about exactly what you found for the earlier part - keep the same character. Attribute a line only to the speaker the passage names. Answer every part, each from the passage that fits it; if a part needs instances from several places, give each one. If no passage plausibly fits a part, say briefly that it was not found, for that part only.

Answer in a single paragraph of about 120 words or fewer, not counting bracketed citations. No headings or bullet points.
\end{quote}

\end{document}